\documentclass[runningheads]{llncs}
\usepackage{graphicx}
\usepackage{comment}
\usepackage{caption}

\usepackage{amsmath,amssymb} 
\usepackage{color}
\usepackage{multirow}
\usepackage{subfigure}

\usepackage[super]{nth}
\newcommand{\tabincell}[2]{\begin{tabular}{@{}#1@{}}#2\end{tabular}}  

\begin{document}

\mainmatter
\def\ECCVSubNumber{2273}  

\title{Balanced Alignment for Face Recognition: A Joint Learning Approach} 

\titlerunning{Abbreviated paper title}
%
\author{Huawei Wei \and
Peng Lu \and
Yichen Wei}
\authorrunning{Wei et al.}
%
\institute{Megvii Technology, Fudan University \\
\email{{weihuawei, weiyichen}@megvii.com}, \email{penglu97@gmail.com}}
\maketitle

\begin{abstract}

Face alignment is crucial for face recognition and has been widely adopted. However, current practice is too simple and under-explored. There lacks an understanding of how important face alignment is and how it should be performed, for recognition. This work studies these problems and makes two contributions. First, it provides an in-depth and quantitative study of how alignment strength affects recognition accuracy. Our results show that excessive alignment is harmful and an optimal balanced point of alignment is in need. To strike the balance, our second contribution is a novel joint learning approach where alignment learning is controllable with respect to its strength and driven by recognition. 
Our proposed method is validated by comprehensive experiments on several benchmarks, especially the challenging ones with large pose.

\end{abstract}

\section{Introduction}

Face alignment is the process that deforms different face images such that their semantic facial landmarks/regions are spatially aligned\footnote{In this work, we only discuss the alignment of 2D images, not 3D face models}. It reduces the geometric variations of faces and eases the face recognition. It is widely used to boost the performance of face recognition, actually serving as an inseparable step.

In spite of its widely perceived effectiveness, in the current research literature, there lacks quantitative analysis about how face alignment affects recognition accuracy, and how face alignment should be performed to aid recognition. 

Existing face recognition approaches exploit alignment in two different and loose ways, which are also barely related. The first localizes facial landmarks via supervised learning and applies a hand-crafted deformation function (\textit{e.g.} affine or projection transformations) to deform the landmarks to match a pre-defined template~\cite{deepid,deepface}. The whole process is before, and unrelated to recognition. The second integrates the image deformation process into the deep neural networks, as pioneered by the Spatial Transformer Networks (STN)~\cite{jaderberg2015spatial} work. Learning of alignment is \emph{unsupervised} and totally driven by the goal of recognition~\cite{wu2017recursive,zhou2018gridface}. There is neither control, nor understanding of how well alignment is performed.

\begin{figure}

\centering{}%

\begin{tabular}{cc}
\vspace{0.1mm}\includegraphics[width=6cm]{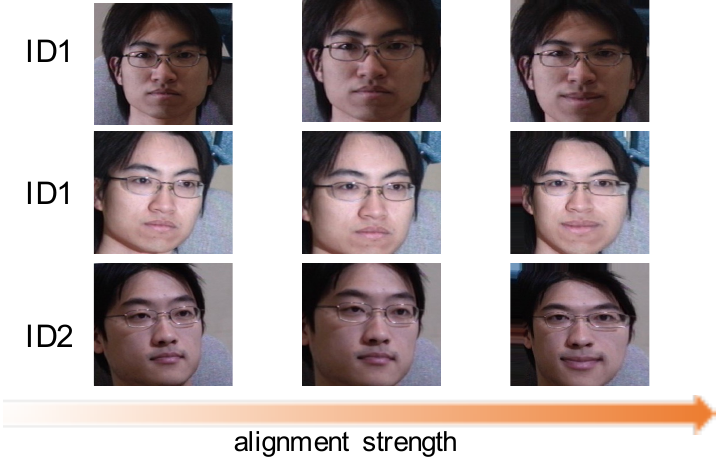} & \includegraphics[width=6cm]{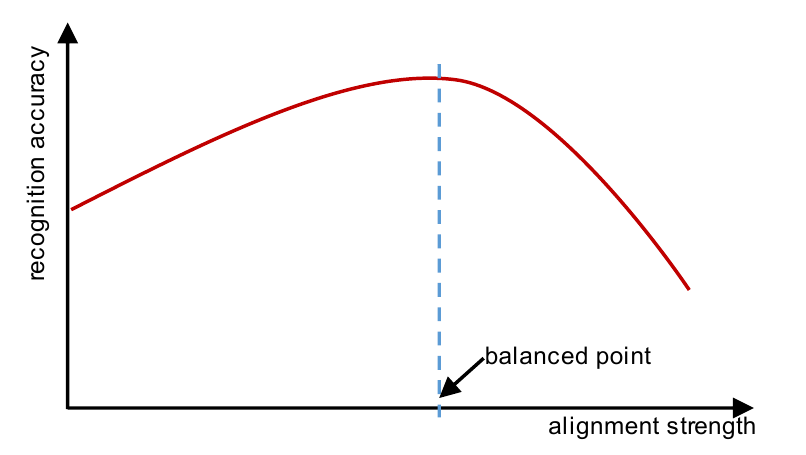}\tabularnewline
(a) & (b)\tabularnewline
\end{tabular}
\caption{\label{fig:align_methods} (a) Visualization of effect of alignment with different strength. The first column is original face images, the second is faces after slight alignment, the third column is faces after excessive alignment (b) Relationship between recognition accuracy and alignment strength. Slight alignment eliminates the pose variation among face images, which makes it easier for recognition. But excessive alignment changes the intrinsic geometries of the face such as the shape of eyes, leading to the loss of identity information. Accordingly, the recognition is confused and the accuracy drops dramatically.}
\vspace{-4mm}
\end{figure}

For the first time, this work performs an in-depth quantitative analysis of how alignment affects recognition and proposes a new learning method, accordingly to the analysis. Our first contribution is a series of comparative experiments that investigate the relationship between alignment strength and recognition accuracy. It is observed that when the strength of alignment increases, the recognition accuracy increases at first and then decreases. This means that \emph{there exists an optimal and balanced point of alignment for the sake of recognition}. This phenomenon is exemplified in Fig.~\ref{fig:align_methods}. A reasonable hypothesis is that, when alignment is moderate, the identity independent geometric variation (in-plane rotation) is eliminated on faces with the same identity. This is helpful for recognition. However, when alignment is excessive, the subtle but crucial facial details/features (e.g., the shape of mouth and face contour) are also attenuated. This compromises recognition accuracy. Our further analysis shows that such compromise brought by the alignment is relieved on feature maps instead of face images.

The observation above indicates that, in order to achieve the optimal tradeoff, \emph{alignment learning should be controllable with respect to its strength and driven by recognition}. No existing approach is adequate for this goal. The second contribution of this work is a novel approach that learns recognition and alignment jointly, while alignment is learned in an auxiliary and adaptive manner. Our approach is based on Spatial Transformer Network (STN)~\cite{jaderberg2015spatial} and equipped with several novel extensions. Alignment is performed on feature maps. Explicit landmark supervision is adopted. Adaptive weights of landmarks are learned. The landmark template is automatically learned, instead of manually designed. All these extensions are validated to improve recognition accuracy.

Our approach is validated by comprehensive experiments on several face recognition benchmarks, especially those challenging for alignment (e.g., with large poses). Solid ablation studies and in-depth analysis are provided.

\section{Related Works}
In this section, we discuss the existing practice of applying face alignment for recognition. There are two categories: face alignment in preprocessing and face alignment in learning. The former consists of 2D/3D transformation relying on a predefined template. The latter is generally based on a deformation network (STN), which can learn spatial transformation from data and can be inserted into recognition system without extra supervision.

\paragraph{\textbf{In Preprocessing}}
The most commonly used face alignment method is using 2D transformation (e.g., affine transformation) to calibrate face images to a predefined template with several landmarks~\cite{deepid}. Another widely used approach is 3D alignment, which fits 3D face model and frontalizes profile faces into canonical view~\cite{deepface,facefront,banerjee2018frontalize}. 3D alignment usually relies on dense landmarks detection or face reconstruction technology~\cite{3ddfa,wei20193d}. 
Compared with these face alignment methods in preprocessing, our proposed alignment scheme does not rely on manual designed template and can be trained end-to-end.

\paragraph{\textbf{In Learning}}
Jaderberg \textit{et al.}~\cite{jaderberg2015spatial} propose Spatial Transformer Networks (STN) to learn invariance to geometric warping (e.g., translation and rotation). STN can be inserted into deep network architecture and optimized in an end-to-end manner. Motivated by the differentiability of STN, Zhong \textit{et al.}~\cite{zhong2017toward} employ STN as an alignment module to automatically learn recognition-oriented spatial transformation for each face image. Wu \textit{et al.}~\cite{wu2017recursive} develop a recursive spatial transformer to learn complex geometric transformation. Lin \textit{et al.}
~\cite{lin2017inverse} connect the Lucas \& Kanade algorithm~\cite{lucas1981iterative} with STN to eliminate geometric variations and gain superior alignment and classification results. To enhance the deformation ability of STN, Zhou \textit{et al.}~\cite{zhou2018gridface} propose a novel non-rigid face rectification method by local
homography transformations. Compared to these STN based face alignment methods, our approach explicitly utilizes a landmarks supervision loss to constrain the optimization of the deformation network. By adjusting the loss ratio, we can control the alignment strength to gain larger benefit. 

\section{In-depth Analysis of Face Alignment for Recognition\label{effect_of_align}}
For face recognition, the importance of face alignment and how it should be performed is rarely discussed in the literature. In this section, we study this question from two perspectives. First, to figure out what alignment is more helpful for recognition, we conduct a deep comparative experiment of several commonly used face alignment methods. The experimental result reveals that excessive alignment (output faces are excessively aligned) is harmful for recognition. 
It is best to align faces at a moderate level.  Second, we analyze how alignment should be conducted. We point out performing alignment on feature maps instead of input images (the way of existing methods) is more advisable. This is due to that alignment on feature map can suppress negative effects of geometric distortion that alignment introduces.

\subsection{Alignment strength influences recognition performance\label{alignment_influence_rec}}
In this part, both qualitative and quantitative analysis of the effect of different face alignment methods on recognition is conducted. Notably, for quantitative analysis, a description indicator to associate different alignment methods is required. So we first give a brief induction about our designed indicator, which is called alignment strength.

Let $\mathcal{S}$ represent the set of landmarks on faces,
where $s\in\mathcal{S}$ represents a specific landmark such as left
eye corner. On face image $I_{i}$,
the coordinate of $s$ is $\boldsymbol{x}_{i}^{s}$ and is transformed
to $\boldsymbol{y}_{i}^{s}$ after deformation. We use Alignment Normalized
Mean Error (ANME) to evaluate the strength of face alignment: 
\begin{equation}
\mbox{ANME}=\frac{1}{N\vert \mathcal S \vert}\sum_{s\in\mathcal{S}}\sum_{i=1}^{N}\frac{\Vert\boldsymbol{y}_{i}^{s}-\bar{\boldsymbol{y}^{s}}\Vert_{2}}{d},\label{eq:ANME}
\end{equation}
where $N$ is the number of samples, $\vert \mathcal{S} \vert$ is the number of landmarks in $\mathcal{S}$, $d$ is the mean inter-pupil distance of all deformed faces, and $\bar{\boldsymbol{y}^{s}}$ is average coordinate of landmark $s$
in deformed images. Intuitively, ANME describes how well faces output by an alignment method are aligned. Low ANME indicates alignment with high strength.

\begin{figure}
	\centering
    	\subfigure[]{
    		\begin{minipage}[b]{0.21\textwidth}
   		 	\includegraphics[width=1\textwidth]{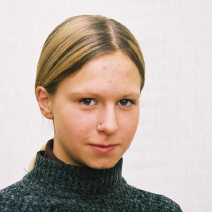}
    		\end{minipage}
		\label{fig:diff-align-ori}
    	}
    	\subfigure[]{
    		\begin{minipage}[b]{0.1\textwidth}
   		 	\includegraphics[width=1\textwidth]{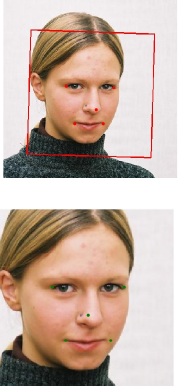}
    		\end{minipage}
		\label{fig:diff-align-affine2d}
    	}
    	\subfigure[]{
    		\begin{minipage}[b]{0.1\textwidth}
   		 	\includegraphics[width=1\textwidth]{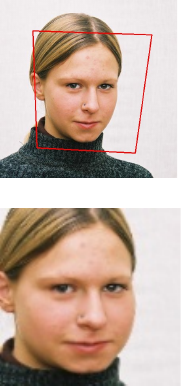}
    		\end{minipage}
		\label{fig:diff-align-stn-proj}
    	}
    	\subfigure[]{
    		\begin{minipage}[b]{0.1\textwidth}
   		 	\includegraphics[width=1\textwidth]{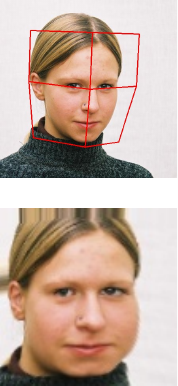}
    		\end{minipage}
		\label{fig:diff-align-stn-tps-grid-2}
    	}
    	\subfigure[]{
    		\begin{minipage}[b]{0.1\textwidth}
   		 	\includegraphics[width=1\textwidth]{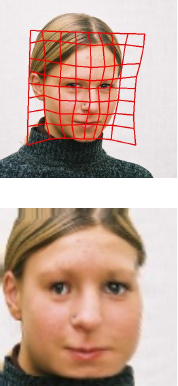}
    		\end{minipage}
		\label{fig:diff-align-tps-grid-8}
    	}
    	\subfigure[]{
    		\begin{minipage}[b]{0.1\textwidth}
   		 	\includegraphics[width=1\textwidth]{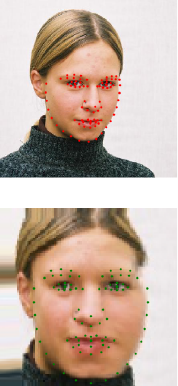}
    		\end{minipage}
		\label{fig:diff-align-full}
    	}
	\caption{Illustration of five alignment methods. (a) Original image; (b) Affine2D; (c) STN-Proj; (d) STN-TPS-Grid-2; (e) STN-TPS-Grid-8; (f) FullAlign. From left to right, the alignment strength increases.}
	\label{fig:diff-align}
\end{figure}

For the study of how alignment affects recognition performance, we select several alignment methods with different strengths for comparison. Specifically, five methods from the commonly used ones are selected, which involve both manually designed and learning-based methods. To be more persuasive, the selected methods have a wide range of alignment strengths. The detailed descriptions of these five methods are as follows:

\textbf{1. Affine2D} is a commonly used simple alignment method in preprocessing of face recognition pipeline~\cite{banerjee2018frontalize}. It aligns images using a 2D affine transformation to match the 5 facial landmarks to a predefined template (i.e., 2D mean face landmarks). Affine2D has  \textit{limited alignment strength} due to its linear form.

\textbf{2$\sim$4. STN-Proj}, \textbf{STN-TPS-Grid-2} and \textbf{STN-TPS-Grid-8} are three variants of STN based alignment methods. They all have stronger deformation ability than Affine2D.
Specifically, STN-Proj uses projective transformation, which has two more degrees of freedom than Affine2D. STN-TPS-Grid-2 and STN-TPS-Grid-8 adopt TPS (thin plate spline)~\cite{bookstein1989principal} transformation, which is a nonrigid and stronger transformation function. Among them, STN-TPS-Grid-8 has larger grid size than STN-TPS-Grid-2, supporting more fine-grained geometric deformation on images. Therefore, STN-TPS-Grid-8 has larger deformation freedom and higher alignment strength. 

\textbf{5. FullAlign} is an extreme alignment method (designed by us) that \textbf{strictly} warps images to a predefined dense landmark template using TPS transformation, where the template is computed using the mean landmarks of frontal face images of training data. FullAlign can obtain \textit{the highest alignment strength} (ANME is 0). Notably, FullAlign is similar but nothing to do with UV mapping in 3D face model~\cite{feng2018joint}. It is a pure operation on 2D space.

An illustration of the above five alignment methods is shown in Fig.~\ref{fig:diff-align}.

\begin{figure}
\centering{}\includegraphics[width=6.5cm]{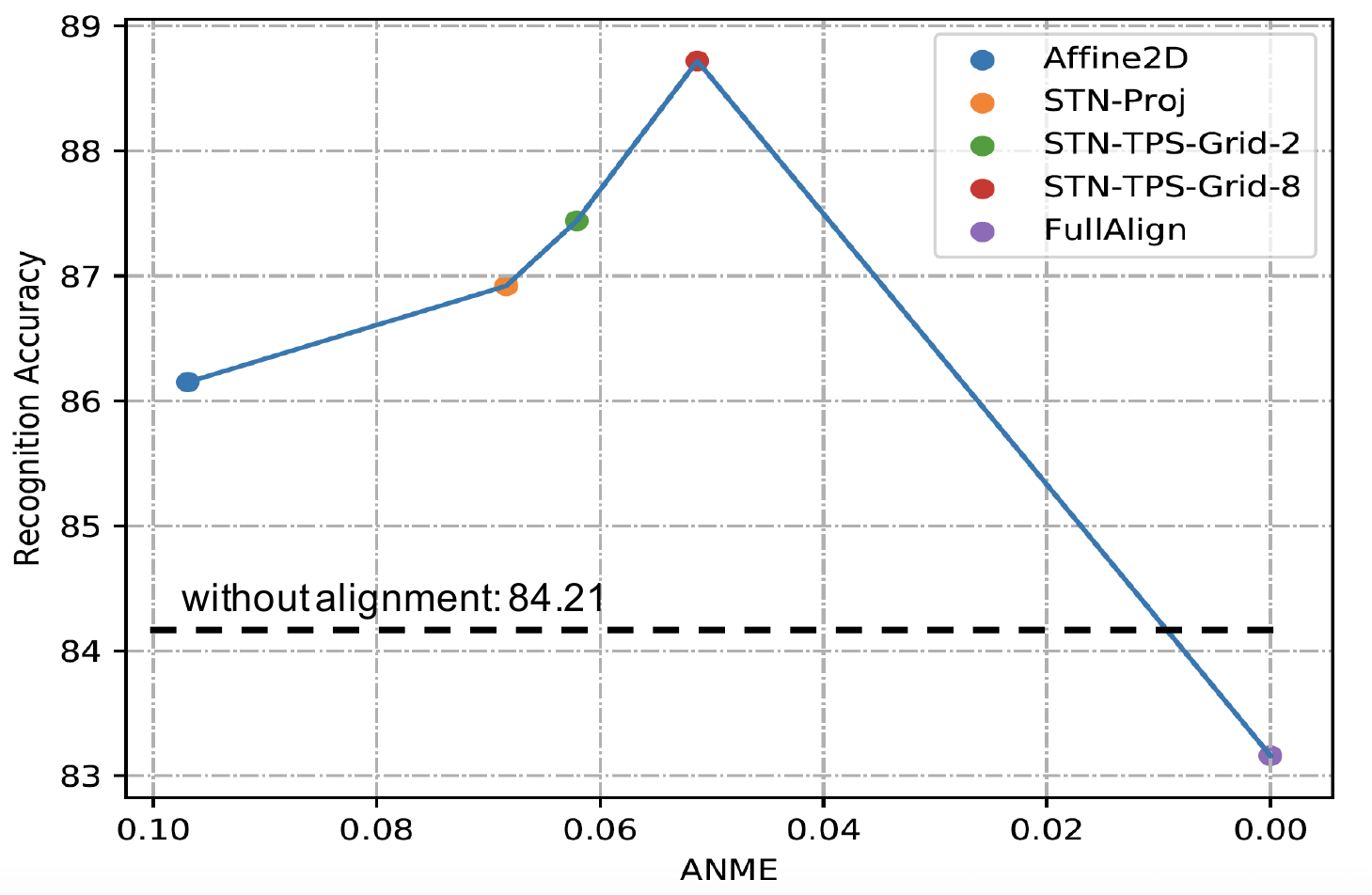}\caption{\label{fig:acc-nme_different_align_methods}Relationship between ANME and recognition performance (rank-1\%) of models using five face alignment methods on MegaFace. As the alignment strength become stronger, the recognition performance increases at first and then decreases.}
\end{figure}

We use ResNet34~\cite{he2016deep} (depicted in Table~\ref{tab:network}) as our recognition network and the commonly used VGGFace2~\cite{cao2018vggface2} as the training dataset. Megaface~\cite{kemelmacher2016megaface}, a challenging and commonly used dataset, is employed as the evaluation benchmark. The other details are the same as that of the experiment in Section~\ref{sec:5-Experiments}. 

Fig.~\ref{fig:acc-nme_different_align_methods} shows the recognition performance and ANME of the abovementioned five alignment methods. 
We notice that when ANME becomes smaller, indicating the gradual increase of alignment strength, the recognition accuracy increases first and then decreases. The results demonstrate that alignment with moderate strength can boost recognition accuracy. But too strong alignment will be harmful for recognition. This phenomenon is intuitively illustrated in Fig.~\ref{fig:align_methods}. When moderate alignment is used (actually it is realized by Affine2D), the identity-independent pose variation in faces is removed (the first column to the second column in Fig.~\ref{fig:align_methods}(a)), thus it makes the recognition easier. When excessive alignment is performed (realized by FullAlign), the intrinsic facial shape, which is crucial for representing a face identity, is distorted (the third column in Fig.~\ref{fig:align_methods}(a)). This will confuse the recognition, so the performance drops dramatically. 

The above experiment demonstrates that alignment reduces the geometric variations of faces but also brings spatial disturbance. The negative effects of the latter will gradually become obvious with the increase of alignment strength. 
To obtain optimal recognition performance, alignment needs to be adjusted to a balanced strength.

Another way to strengthen the benefit of alignment is to relieve the negative effects of its introduced spatial disturbance. In the next subsection, we point out alignment on feature maps can effectively achieve this goal.


\begin{table}[t!]
\begin{centering}
\begin{tabular}{c|c|c|c}
\hline 
{\footnotesize{}Stage Name} & {\footnotesize{}Input Size} & {\footnotesize{}Output Size} & {\footnotesize{}Network Structure}\tabularnewline
\hline 
\hline 
{\footnotesize{}stage0} & {\footnotesize{}112$\times$112} & {\footnotesize{}112$\times$112} & {\footnotesize{}$\left[\begin{array}{cc}
3\times3, & 64\end{array}\right]$}\tabularnewline
\hline
\multirow{1}{*}{{\footnotesize{}stage1}} & \multirow{1}{*}{{\footnotesize{}112$\times$112}} & {\footnotesize{}56$\times$56} & {\footnotesize{}$\left[\begin{array}{cc}
3\times3, & 64\\
3\times3, & 64
\end{array}\right]$x3}\tabularnewline
\hline
{\footnotesize{}stage2} & {\footnotesize{}56$\times$56} & {\footnotesize{}28$\times$28} & {\footnotesize{}$\left[\begin{array}{cc}
3\times3, & 128\\
3\times3, & 128
\end{array}\right]$x4}\tabularnewline
\hline
{\footnotesize{}stage3} & {\footnotesize{}28$\times$28} & {\footnotesize{}14$\times$14} & {\footnotesize{}$\left[\begin{array}{cc}
3\times3, & 256\\
3\times3, & 256
\end{array}\right]$x6}\tabularnewline
\hline
{\footnotesize{}stage4} & {\footnotesize{}14$\times$14} & {\footnotesize{}7$\times$7} & {\footnotesize{}$\left[\begin{array}{cc}
3\times3, & 512\\
3\times3, & 512
\end{array}\right]$x3}\tabularnewline
\hline
{\footnotesize{}output layer} & {\footnotesize{}7$\times$7} & {\footnotesize{}1$\times$1} & \tabincell{c}{\footnotesize{}BN-\textgreater Dropout--\textgreater \\ \footnotesize{}512-d FC--\textgreater \footnotesize{}BN}\tabularnewline
\hline 
\end{tabular}
\par\end{centering}
\centering{}\vspace{2mm}
 \caption{\label{tab:network}Details of the Recognition Network architecture.}
 \vspace{-8mm}
\end{table}

\begin{figure}
\centering{}\hspace{-2mm}\includegraphics[width=8cm, height=5cm]{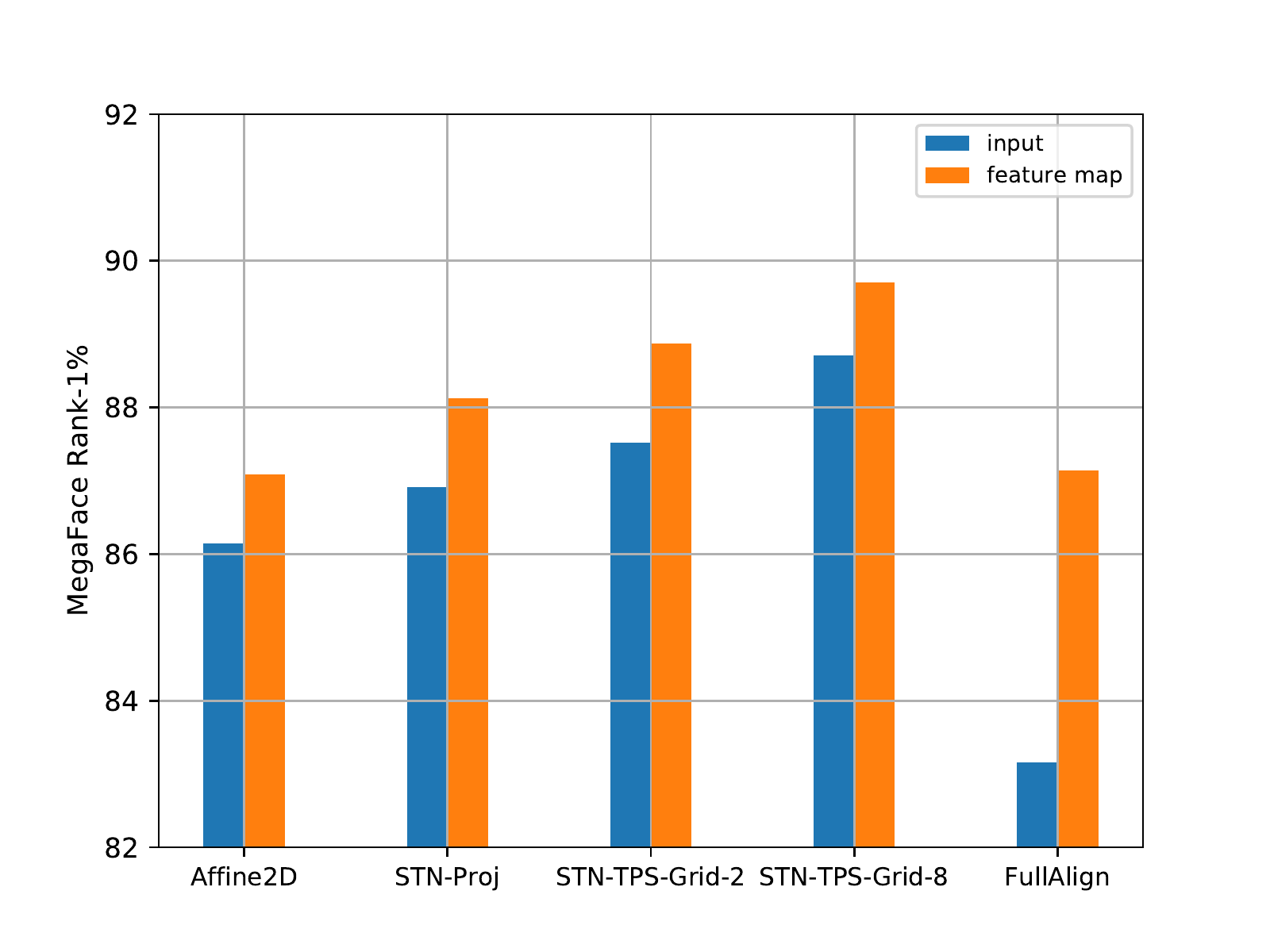}
\vspace{-2mm}
\caption{\label{fig:UV-warp_result}Recognition performance of models with different alignment methods. We report the Rank-1\% on MegaFace. We can find alignment on feature maps always gains better performance than input images}
\end{figure}

\subsection{\label{subsec:face_alignment_on_feature_map}Alignment on feature maps is more robust}
From the previous subsection, we know that alignment introduces geometric distortion, which is harmful to recognition. In this section, we experimentally find that alignment on feature maps can effectively prevent the negative impact of it. Though this idea has been explored by researchers for other tasks~\cite{kanazawa2016warpnet,qi2017pointnet,liu2019hierarchical}, it is used on face alignment and recognition task for the first time. Moreover, it is the first time that an in-depth analysis of its effectiveness is provided.

We use the five alignment methods introduced in Section ~\ref{alignment_influence_rec} to align input images and feature maps respectively. The feature maps are the output of stage 0 of the recognition network (Table~\ref{tab:network}), which have the same resolution as input images.

The result is shown in Fig.~\ref{fig:UV-warp_result}.
We can note that alignment on feature maps always obtains better performance than input images. 

We attribute this phenomenon to the following explanation:
as feature maps are generated by convolution operation,
the local geometric information, which is critical to represent the identity of a face, is encoded into each cell of feature maps. Hence, the local geometric information is not affected by spatial displacement caused by alignment. 
However, since the position relations of pixels in input images are broken after alignment, the local geometric information will be destroyed. Therefore alignment on input images gains worse recognition performance. 

\vspace{-3mm}
\section{\label{subsec:deformation_with_landmark_supervision}Jointly Learning Alignment and Recognition}
\begin{figure}[t!]
\centering{}\includegraphics[width=12cm]{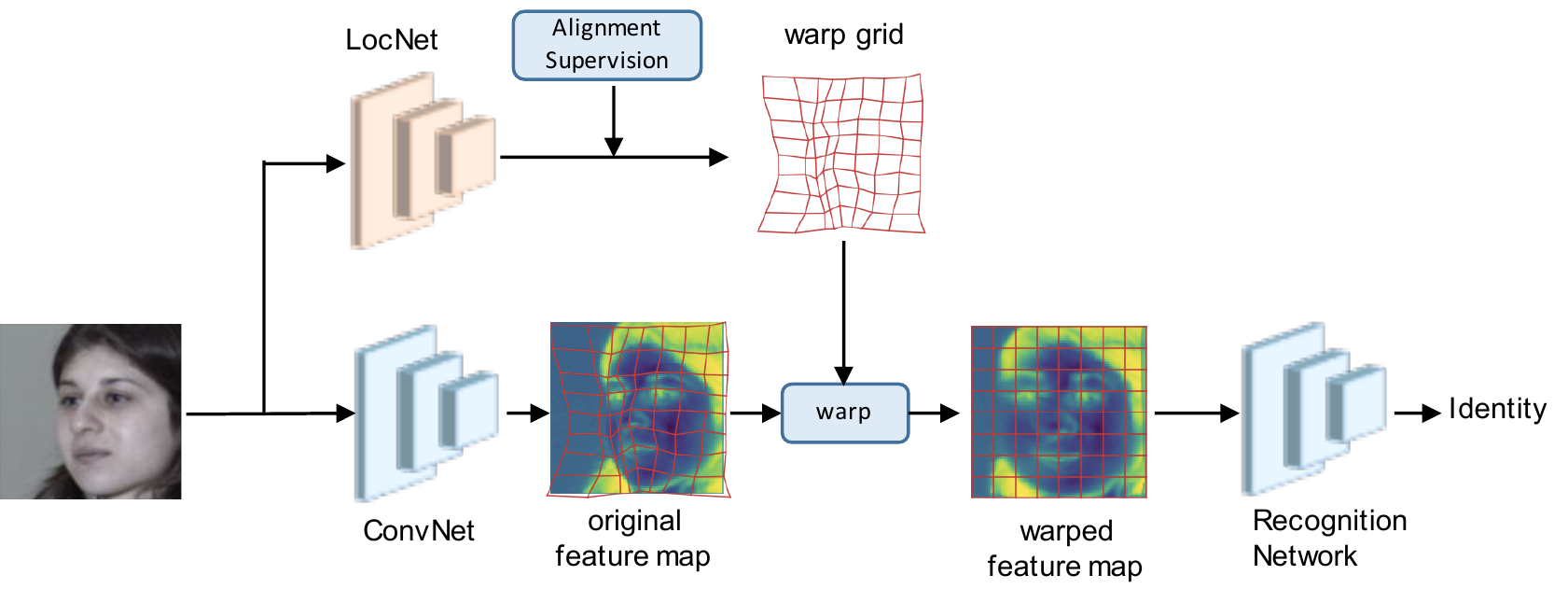}\caption{\label{fig:warp_process}Overview of our approach. A warp grid is first generated by LocNet from the input image, and at the same time the image is sent into ConvNet to generate the feature map. Afterwards, the feature map is warped by the generated warp grid. Finally, the warped feature map is fed into the Recognition Network for face identification. The alignment supervision is added on the LocNet and forces the generated warp grid to minimize the distance between landmarks of each aligned face and a landmark template (more details are shown in Fig.~\ref{fig:Landmark-Supervision-Loss}). }
\end{figure}

\vspace{-2mm}
Section~\ref{alignment_influence_rec} demonstrates that optimal recognition performance can be gained by alignment with balanced strength. A solution is to design an alignment method with the ability to adjust alignment strength. By tuning the strength, the balance can be struck. To our knowledge, all existing methods do not meet this requirement. In this section, we realize the strength controllable alignment by  jointly learning alignment and recognition. The alignment process relies on a deformation network (\textit{i.e.,} Spatial Transformer Network~\cite{jaderberg2015spatial} in our practice) and a explicit landmark alignment loss. The loss constrains all deformed faces to approach a learned landmark template in a soft manner. By adjusting the constraint degree, the balanced alignment strength can be achieved. The overview of our approach is shown in Fig.~\ref{fig:warp_process}

\subsection{Review of Spatial Transformer Network.\label{DeformNet}} 
We first review the deformation network (\textit{i.e.,} Spatial Transformer Network(STN)).
 STN is a differentiable module that enables the network to conduct spatial transformations on images. 
 It consists of three components: 
 
 \noindent 1) \textit{Localisation Network}, which takes an image as input and output a group of transformation parameters. The number of parameters is determined by the degree of freedom of the chosen transformation modes (e.g. 6 for affine transformation and 8 for projective transformation); 
 
 \noindent2) \textit{Grid Generator}, which generate a warp grid using the transformation parameters. STN is compatible with many transformations, including affine transformation, projective transformation, thin plate spline (TPS)~\cite{bookstein1989principal}, and etc. 
 
 \noindent3) \textit{Differentiable Sampler}, which samples pixel values on the input image based on the warp grid and generates the transformed image. In the sampling process, bilinear interpolation is generally utilized. 

In our practice, the Localisation Network is a slim version of Resnet18~\cite{he2016deep}. The number of channels of stage 0 $\sim$ 4 are 8, 16, 32, 64, and 64 respectively. For the Grid Generator, we use TPS as the transformation. TPS requires a set of source points and target points to generate a warp grid. We manually set target points as the cross points of a regular grid, where the grid size is set as 8$\times$8 (see the grid on warped feature map in Fig.~\ref{fig:warp_process}). The source points are obtained through the Localisation Network. In the Differentiable Sampler, we use bilinear interpolation. 

Note that our deformation process is conducted on feature maps. This is based on the conclusion in Section~\ref{subsec:face_alignment_on_feature_map} that feature maps are more robust to geometric distortion introduced by deformation. The deformation process is illustrated in Fig.~\ref{fig:warp_process}. A facial image is used to extract feature maps and compute TPS transformation parameters. Then the generated feature maps is warped by the TPS transformation.

\subsection{Strength controllable alignment learning\label{lmksup}} 
To enable the deformation network to dynamically adjust alignment strength, we develop a facial landmark alignment loss $\mathcal{L}_{\text{align}}$ to supervise the learning of the deformation network. The proposed loss is defined as: 
\vspace{-3mm}

\begin{align}
& \mathcal{L}_{\text{align}}=\mathcal{L}_{\text{lmk}} + \mathcal{L}_{\text{reg}}, \\
& \mathcal{L}_{\text{lmk}}=\frac{1}{N\vert \mathcal{S} \vert}\sum_{i=1}^{N}\sum_{s\in\mathcal{S}}\alpha^{s}\Vert W(\boldsymbol{T}^{s}; I_i)-\boldsymbol{x}_{i}^{s}\Vert_{2}\label{eq:3.3-lmk loss}, \\
& \mathcal{L}_{\text{reg}}=\sum_{s\in\mathcal{S}}\left(1-\alpha^{s}\right)^{2}.\label{eq:3.3-regularization-loss} 
\end{align}

\noindent In Eq.~\ref{eq:3.3-lmk loss} , $\alpha^{s}$ is the loss weight for each individual facial landamark $s$. It is a learnable variable and constrained to (0, 1) by sigmoid function in training procedure.
$\boldsymbol{T}^{s}$ is one specific point of the facial landmark template $\boldsymbol{T}=\{\boldsymbol{T}^{s}|s\in\mathcal{S}\}$. It is also a variable that can be optimized towards recognition.
$W$ represents the deformation process of warping $\boldsymbol{T}^{s}$ to its transformed version.  Eq.~\ref{eq:3.3-regularization-loss} is a regularization that prevents $\alpha^{s}$ to be optimized to 0.

\begin{figure}[t!]
\begin{centering}
\hspace{-5mm}
\includegraphics[width=8cm]{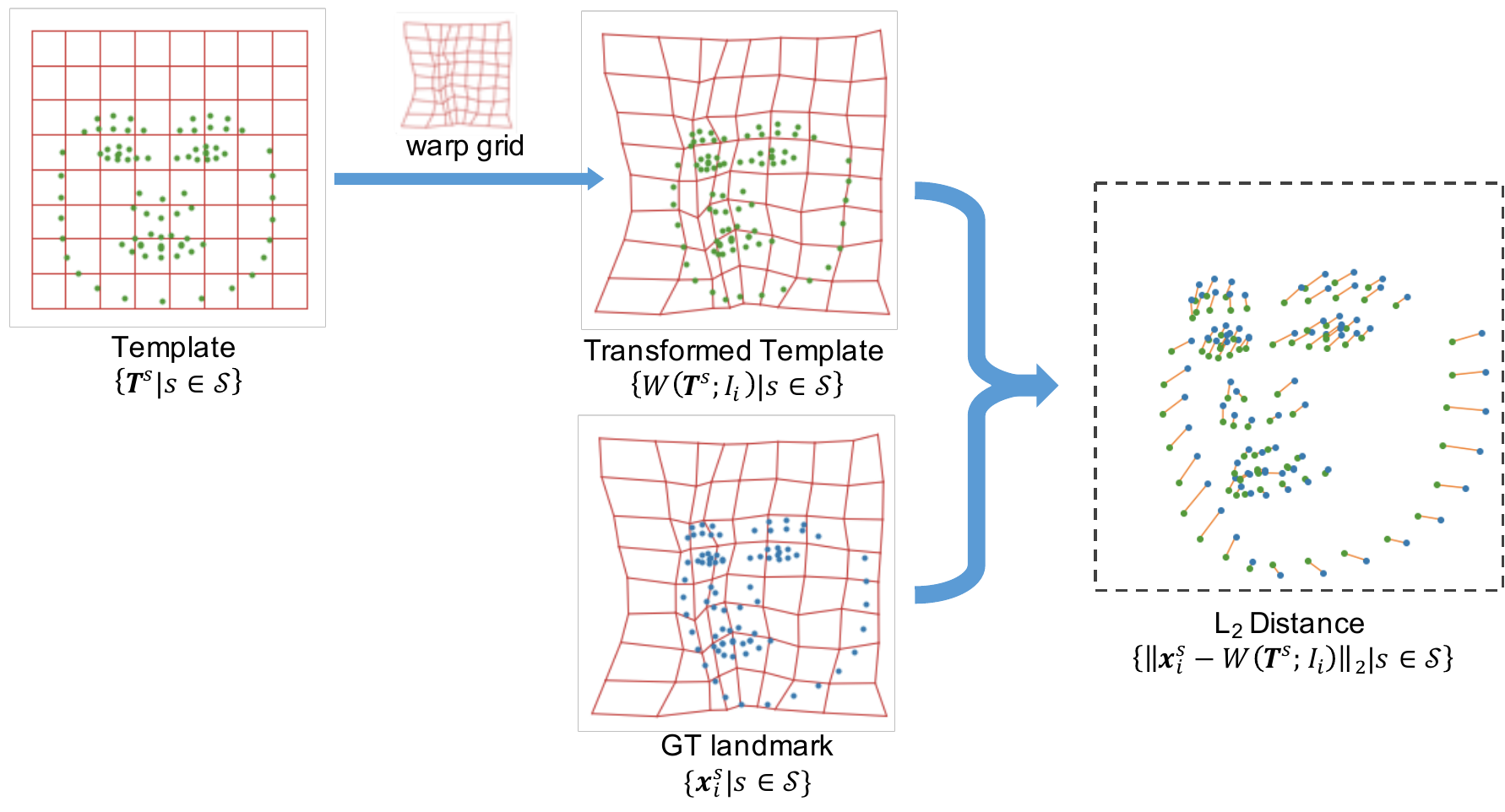}\vspace{2mm}
\par\end{centering}
\centering{}\caption{\label{fig:Landmark-Supervision-Loss}Visualization of the computation of alignment supervision loss $\mathcal{L}_{\text{align}}$. The TPS transformation, which is illustrated by the warp grid, is generated from $I_i$ with LocNet (see Fig.~\ref{fig:warp_process}). The function of this loss is to supervise the LocNet to learn a warp grid that makes all deformed faces close to the template. }
\vspace{-3mm}
\end{figure}

Intuitively, by minimizing $\mathcal{L}_{\text{align}}$, all faces are aligned to the template $\boldsymbol{T}$ as much as possible, which leads to lower ANME and alignment with high strength. An illustration of $\mathcal{L}_{\text{align}}$ is shown in Fig.~\ref{fig:Landmark-Supervision-Loss}.

To give a clearer idea of why $\mathcal{L}_{\text{align}}$ is designed in this way, we analyze each factor of it in detail:

\noindent \textbf{Individual landmark loss weight $\alpha^{s}$.} The alignment of different facial regions has different effects on recognition. For example, contours that describe the shape of a face cannot be strictly aligned. Otherwise, it will cause severe geometric distortion.
To avoid this problem, we introduce $\alpha^{s}$ to enable $\mathcal{L}_{\text{align}}$ to impose different alignment constraints on different facial regions. $\alpha^{s}$ can be adaptively optimized toward face recognition.

\noindent \textbf{Learnable template $\boldsymbol{T}$.} The template $\boldsymbol{T}$ serves as a center of landmarks for all deformed faces. That is, all faces are aligned to this template as much as possible after deformation. The template is a learnable variable that can be optimized in a recognition oriented manner. By this design we can get rid of manual design of landmark template, which may not be optimal for recognition.

\noindent \textbf{The regularization loss $\mathcal{L}_{\text{reg}}$.} The distance between ground truth landmark $\boldsymbol{x}_{i}^{s}$ and the ``predicted landmark'' $W(\boldsymbol{T}^{s}; I_i)$ is non-negative. Without the regularization item, simply minizing $\mathcal{L}_{\text{align}}$ will make the loss reduce to 0 by assigning all landmark weight $\alpha^s$ with 0. $\mathcal{L}_{\text{reg}}$ can effectively prevent the vanishment of $\mathcal{L}_{\text{align}}$.

The complete loss function to train our model is: 
\begin{equation}
\mathcal{L}=\mathcal{L}_{\text{fr}}+\lambda\mathcal{L}_{\text{align}},\label{eq:3.3-total-loss}
\end{equation}
 $\mathcal{L}_{\text{fr}}$ is the face recogniton classification loss. We use AM-Softmax~\cite{wang2018additive} with margin 0.35 and scale 64 as $\mathcal{L}_{\text{fr}}$.
$\lambda$ is a hyper-parameter controlling the degree of the alignment supervision constraint.

The parameter $\lambda$ provides $\mathcal{L}_{\text{align}}$ with the ability to adjust the strength of alignment.
A large $\lambda$ leads to strong alignment and vice versa. By tuning the value of $\lambda$, we can adjust the alignment strength to the balanced point and thus achieve superior recognition performance. In our practice, we find the optimal value of $\lambda$ is 3. And we use $\lambda=3$ by default in all experiments.

\vspace{-1mm}
\section{\label{sec:5-Experiments}Experiments}

\subsection{Datasets and implementation details}
\noindent \textbf{Datasets}.
Our models are learned from a large public dataset - VGGFace2~\cite{cao2018vggface2}, which contains 3.3+ million images. 
We evaluate our models on several public benchmarks:  MegaFace~\cite{kemelmacher2016megaface}, LFW~\cite{huang2008labeled}, YTF~\cite{wolf2011face} and Multi-PIE~\cite{gross2010multi}. Multi-PIE contains 754,200 images from 337 subjects. The images vary in pose, illumination, and expression. A typical protocol~\cite{yim2015rotating} which selects only 137 subjects for testing is already saturated and it is difficult to distinguish the effectiveness of our method. So we use full 337 subjects for testing. 
 The MegaFace dataset we utilize is refined using the same protocol as in ArcFace~\cite{deng2019arcface}.

\noindent \textbf{Implementation Details}.
We use Resnet34 as the recognition backbone if not specified.
The input images are resized to 112$\times$112. The RGB value of each pixel is normalized to [-1, 1].  
We use SGD~\cite{robbins1951stochastic} optimizer with momentum 0.9 and the weight decay is 5e-4. The batch size is 256 and we train our model for 24 epochs. The learning rate is set to 0.2 at the beginning of training and divided by 10 at the beginning of the \nth{13}, \nth{18}, and \nth{23} epoch. 
For the alignment supervision loss, we extract 84 key points using models in~\cite{3ddfa}.  All experiments are implemented by PyTorch~\cite{paszke2017automatic} on NVIDIA GeForce RTX 2080 Ti GPUs.

\begin{figure}[t!]
\centering{}%
\begin{tabular}{cc}
\includegraphics[width=5cm]{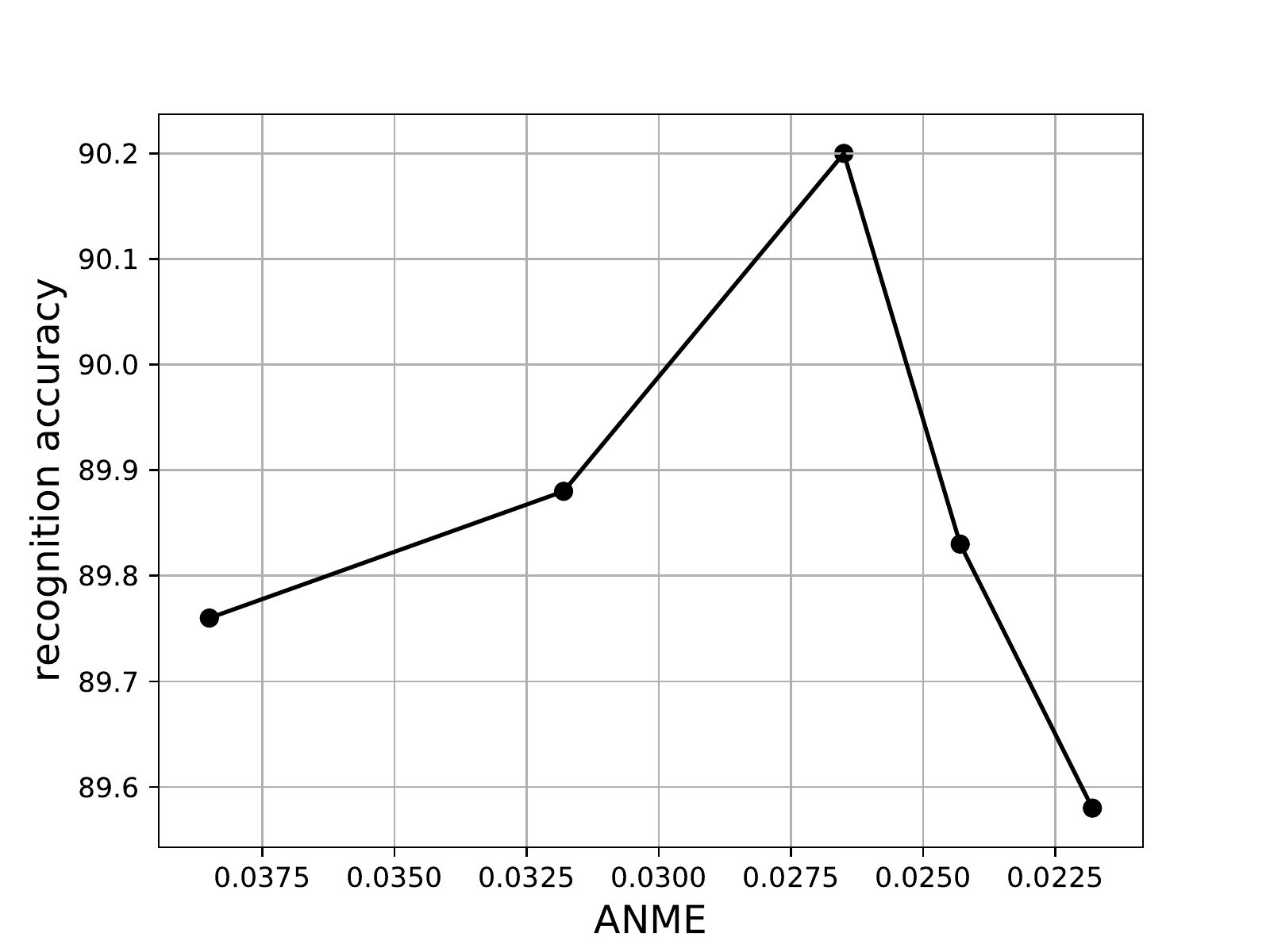} & \includegraphics[width=5cm]{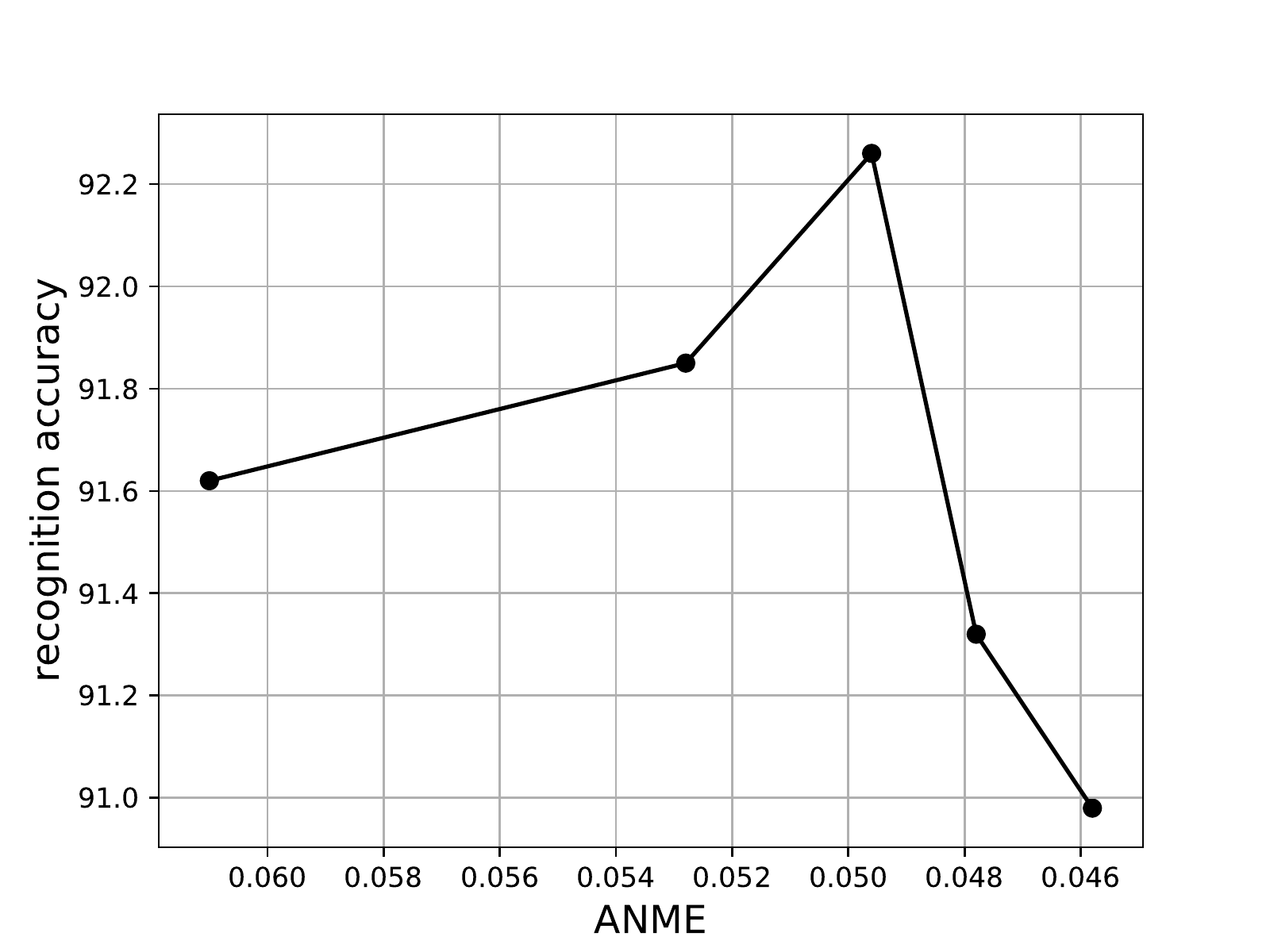}\tabularnewline
(a) results on MegaFace & (b) results on Multi-PIE\tabularnewline
\end{tabular}\caption{\label{fig:loss-ratio}The illustration of how $\lambda$ controls alignment strength and influences recognition performance. For each subfigure, the five scatter points from left to right represent the results of $\lambda =\{0,1,3,5,7\}$. We can see that as $\lambda$ gets larger, the ANME becomes smaller, which represents the enhancement of alignment strength. The recognition accuracy increases at first and then decreases, the optimal accuracy is obtained at the balanced point where $\lambda$=3.}
\end{figure}

\subsection{Effect of balanced alignment learning}

In our approach, the loss ratio $\lambda$ controls the alignment strength, which further affects the recognition performance. To verify that the balanced alignment can be achieved by tuning $\lambda$, we conduct extensive experiments with different $\lambda$ on MegaFace and Multi-PIE datasets. Specifically, $\lambda$ is set as \{0, 1, 3, 5, 7\}.
Note that it is actually the deterministic STN based alignment method (without alignment supervision $\mathcal{L}_{\text{align}}$) when $\lambda=0$. It is considered as our baseline in our following experiments.

The results are depicted in Fig.~\ref{fig:loss-ratio}.
We can see that ANME decreases as $\lambda$ increases, which verifies increasing $\lambda$ enhances the alignment strength. As the alignment gets stronger, the recognition performance increases at first and then decreases. This phenomenon demonstrates the existence of a balanced point of alignment strength for recognition performance. By fine adjustment of the alignment strength, superior performance can be gained. In our experiment, when $\lambda$ is taken as 3, the best performance is obtained.

\subsection{Ablation experiments of our approach}
\subsubsection{Effect of alignment on feature maps\label{subsec:analysis-fmap}}
From Section~\ref{subsec:face_alignment_on_feature_map}, we know that feature maps are more robust to geometric distortion introduced by alignment. Therefore, performing alignment on feature maps can gain larger benefit. We give a further verification in this part.

The compared models are as follows:

\noindent \raisebox{0.5mm}{---}\textit{baseline, align@input/align@fmap}: the deterministic STN based face alignment method, the alignment is performed on input images or feature maps of the stage 0 in the recognition network.

\noindent \raisebox{0.5mm}{---}\textit{our approach, align@input/align@fmap}: based on the baseline model, $\mathcal{L}_{\text{align}}$ is utilized to supervise the training of the deformation network.

\begin{table}[t!]
\begin{centering}
\begin{tabular}{c|cccccc|c}
\hline 
\multirow{2}{*}{{Settings}} & \multicolumn{6}{c|}{{Multi-PIE}} & \multirow{2}{*}{{MegaFace}} \tabularnewline
\cline{2-7} \cline{3-7} \cline{4-7} \cline{5-7} \cline{6-7} \cline{7-7} 
 & {$90^{\circ}$ } & {$75^{\circ}$ } & {$60^{\circ}$ } & {$45^{\circ}$ } & {$30^{\circ}$ } & {$15^{\circ}$ } 
 \tabularnewline
\hline 
\hline 
{baseline, align@input} & {74.71 } & {86.66 } & {92.85 } & {95.78} & {96.97 } & {97.41 } & {88.71 } \tabularnewline
{baseline, align@fmap } & {76.87 } & {88.20 } & {93.70 } & {96.20 } & {97.19 } & {97.53 } & {89.71 } \tabularnewline
\hline 
{our approach, align@input} & {76.20 } & {88.04 } & {93.82 } & {96.29 } & {97.22 } & {97.62 } & {89.98} \tabularnewline
{our approach, align@fmap } & \textbf{78.06}{ } & \textbf{89.05}{ } & \textbf{94.38}{ } & \textbf{96.66}{ } & \textbf{97.51}{ } & \textbf{97.88}{ } & \textbf{90.20}{ } \tabularnewline
\hline 
\end{tabular}
\par\end{centering}
\centering{}\vspace{2mm}
 \caption{\label{tab:Recognition-results}Recognition results with different
align settings. We evaluate the recognition result by rank-1\%.}
\end{table}

\begin{figure}[t!]
\vspace{-2mm}
\centering{}%
\begin{tabular}{cc}
\includegraphics[width=5cm]{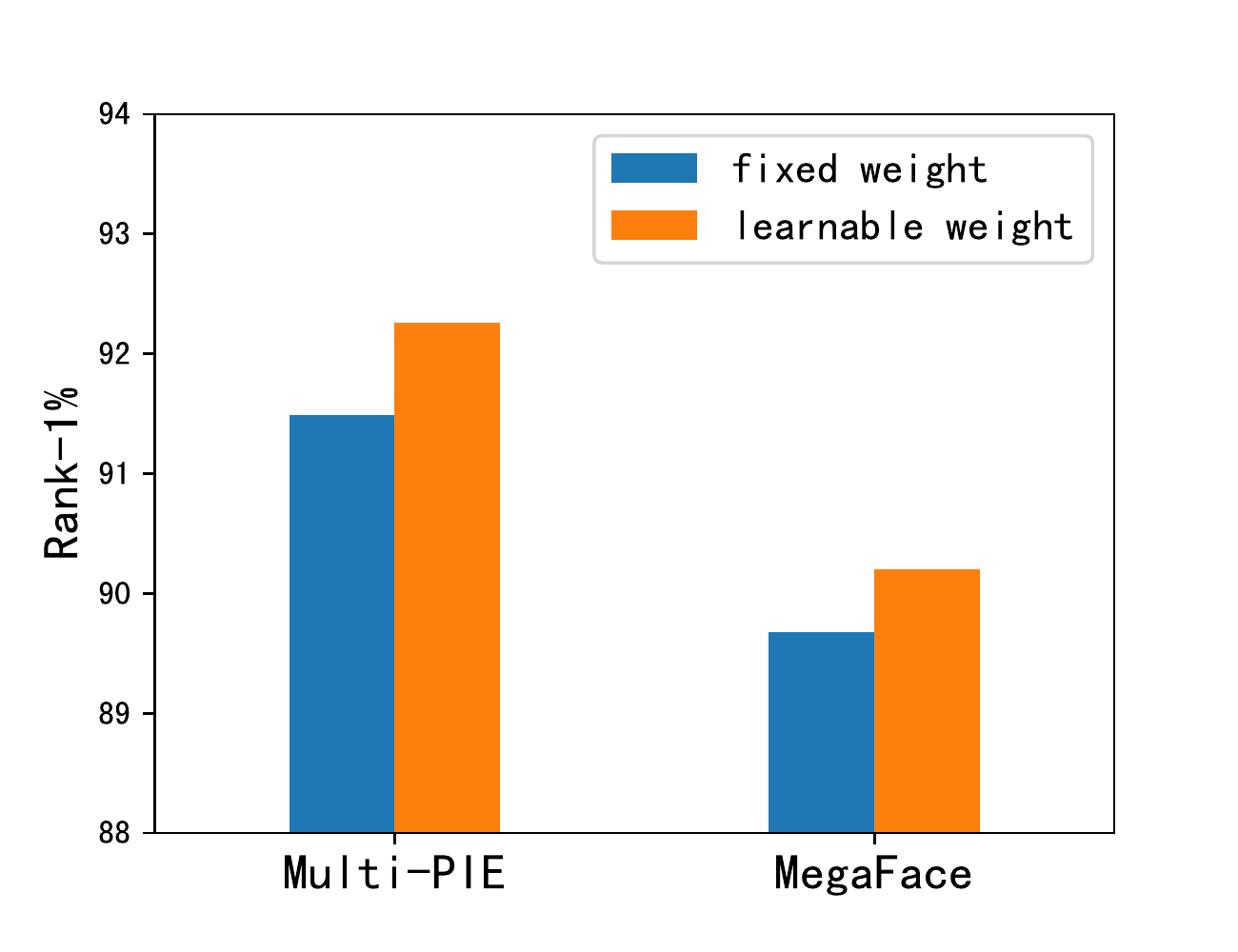} & \includegraphics[width=5cm]{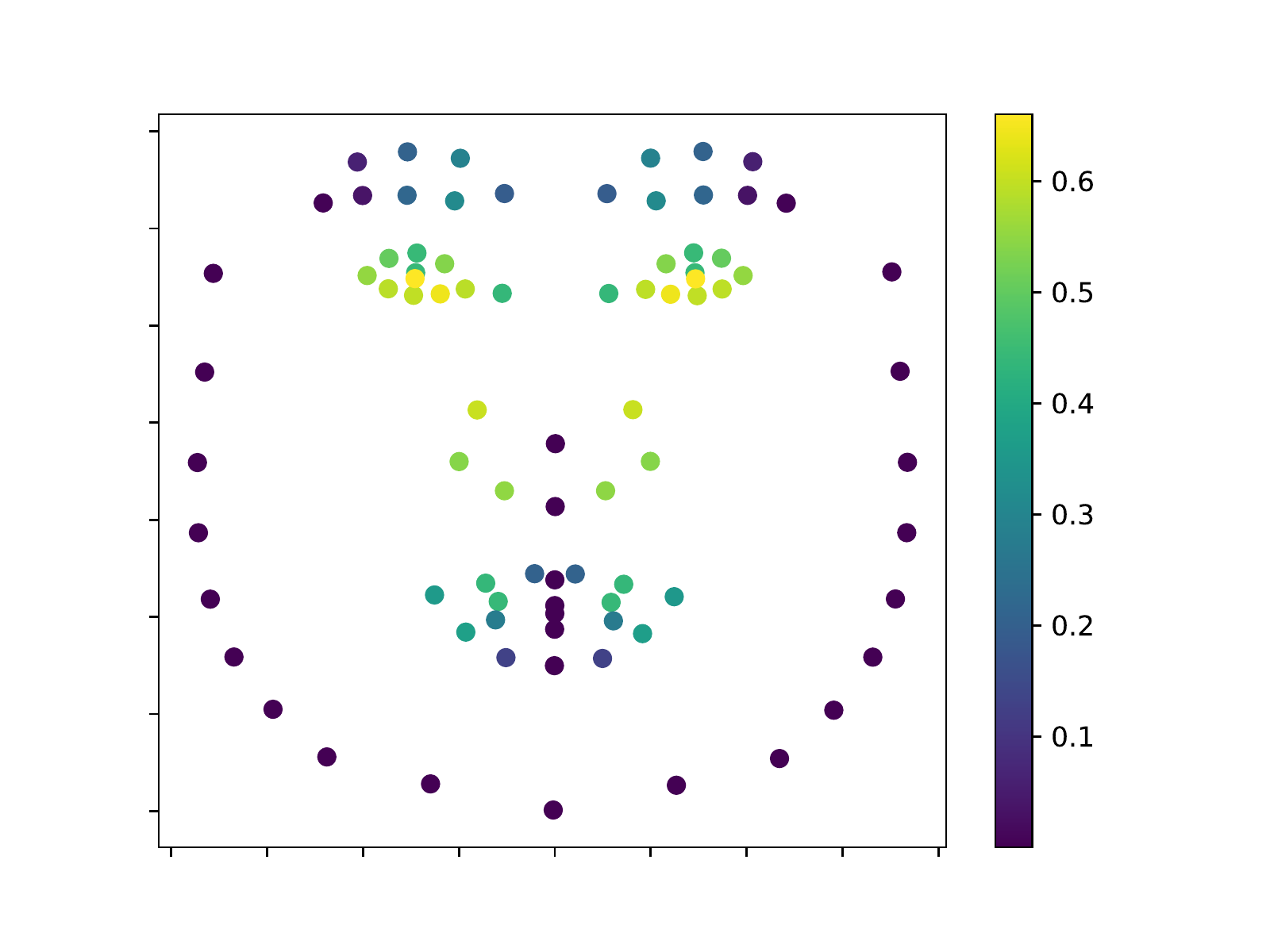}\tabularnewline
(a) & (b)\tabularnewline
\end{tabular}\caption{\label{fig:4.4.3-lmk weight}(a) Recognition accuracy of models trained
with fixed/learnable landmark weight on Multi-PIE and MegaFace. We
report rank-1 and the result on Multi-PIE is averaged among all poses.
(b) Our learned landmark weights. Dark color means low landmark weight
and vice versa. }
\end{figure}

The results is shown in in Table~\ref{tab:Recognition-results}.
It can be noticed that either the baseline or our approach, alignment on feature maps always achieves better performance than input images.
 To be specific, 2.16\%/1.86\% improvements on $90^{\circ}$  case of Multi-PIE is gained in baseline/our approach, and 1.00\%/0.22\% improvements is achieved on MegaFace. It reveals that the recognition does benefits from alignment on feature maps.

\vspace{-4mm}
\subsubsection{Effect of landmark loss weight $\alpha^{s}$ \label{subsec:lmk_weight}}
As discussed in Section~\ref{lmksup}, $\alpha$ represents the alignment strength of different regions of a face. Compared with manually setting these parameters to a constant, dynamically adjusting them in the learning procedure enables the LocNet to assign different significance on different face regions based on their intrinsic impact on recognition. Fig.~\ref{fig:4.4.3-lmk weight} (a) shows that this scheme improves the recognition accuracy on two benchmarks. 

To figure out the effect of alignment of different parts of human face on recognition, we show the learned $\alpha$ of our approach in Fig.~\ref{fig:4.4.3-lmk weight} (b). As can be seen, the weights of contour and middle regions of the face are very low. This means that these areas cannot be strongly aligned. The reason may be that the shape of these regions largely represent the intrinsic shape of the face. When these positions are strongly aligned, serious geometric distortion will be introduced.

\begin{table}[t!]
\setlength{\belowcaptionskip}{-4mm}  
\begin{centering}
\begin{tabular}{c|cccccc|c}
\hline 
\multirow{2}{*}{{Settings}} & \multicolumn{6}{c|}{{Multi-PIE}} & \multirow{2}{*}{{MegaFace}}\tabularnewline
\cline{2-7} \cline{3-7} \cline{4-7} \cline{5-7} \cline{6-7} \cline{7-7} 
 & {$90^{\circ}$ } & {$75^{\circ}$ } & {$60^{\circ}$ } & {$45^{\circ}$ } & {$30^{\circ}$ } & {$15^{\circ}$ } & \tabularnewline
\hline 
\hline 
{align@input, with $\boldsymbol{T_{fix}}$} & {76.32 } & {87.72 } & {93.59 } & {96.12 } & {97.09 } & {97.49 } & {89.22}\tabularnewline
{align@input, with $\boldsymbol{T_{fix}}^{\dagger}$} & {\textbf{76.30} } & {\textbf{88.18} } & {\textbf{93.86} } & {\textbf{96.33} } & {\textbf{97.24} } & {97.60 } & {89.72}\tabularnewline
{align@input, with $\boldsymbol{T}$} & {76.20 } & {88.04 } & {93.82 } & {96.29 } & {97.22 } & {\textbf{97.62} } & {\textbf{89.98}}\tabularnewline
\hline 
{align@fmap, with $\boldsymbol{T_{fix}}$} & {76.32 } & {88.08 } & {94.00} & {96.41 } & {97.36 } & {97.76 } & {89.60 }\tabularnewline
{align@fmap, with $\boldsymbol{T_{fix}}^{\dagger}$} & {\textbf{78.07} } & {\textbf{89.23} } & {\textbf{94.42}} & {96.65 } & {97.50 } & {97.83 } & {\textbf{90.37} }\tabularnewline
{align@fmap, with $\boldsymbol{T}$} & {78.06 } & {89.05 } & {94.38} & {\textbf{96.66} } & {\textbf{97.51} } & {\textbf{97.88 }} & {90.20}\tabularnewline
\hline 
\end{tabular}
\par\end{centering}
\centering{}\vspace{2mm}
 \caption{\label{tab:different template}Recognition results with fixed template and learnable template. $^{\dagger}$ means we use the learned template (Fig.~\ref{fig:diff-template} (b)) instead of pre-computed template (Fig.~\ref{fig:diff-template} (a)), as the fixed template to supervise the training of deformation network. 
We evaluate the recognition result by rank-1\%. }
\end{table}

\begin{figure}[t!]
\setlength{\belowcaptionskip}{-4mm}  
\centering{}%
\begin{tabular}{cc}
\includegraphics[width=4.5cm]{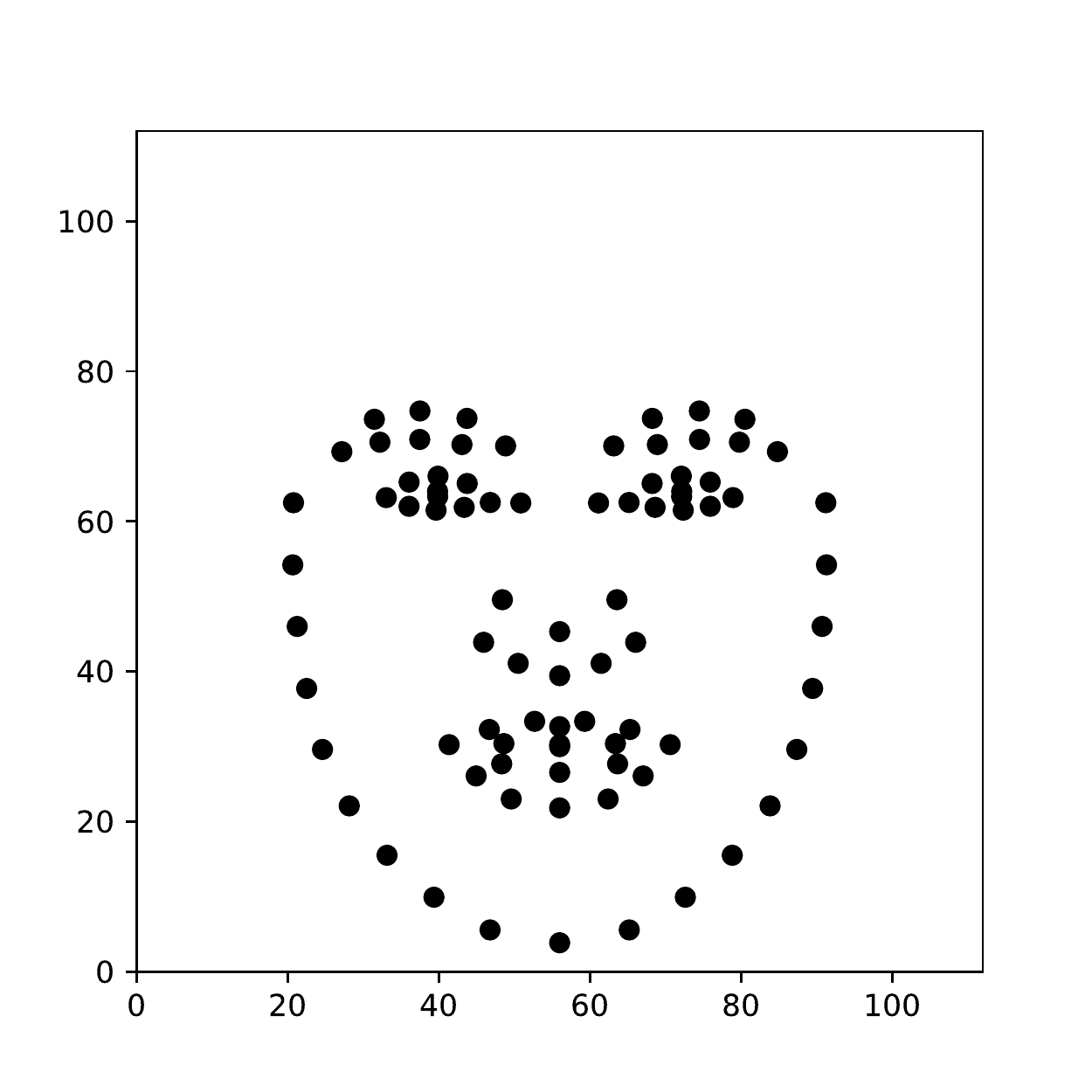} & \includegraphics[width=4.5cm]{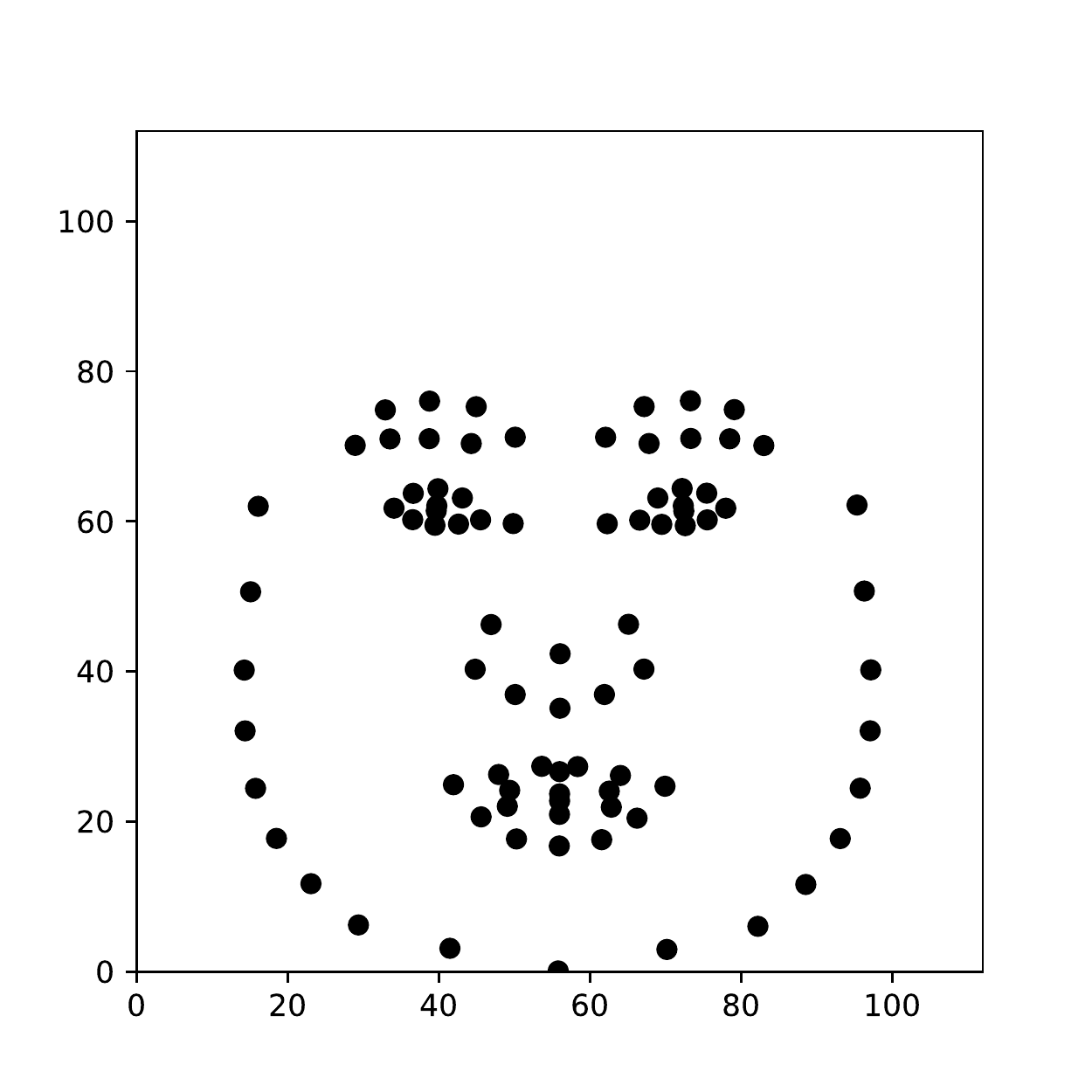}\tabularnewline
(a) fixed template & (b) learnable template\tabularnewline
\end{tabular}\caption{\label{fig:diff-template} Illustration of fixed template and learnable
template. Landmarks in the learnable template is relatively loose
than that of the fixed template.}
\end{figure}

\vspace{-3mm}
\subsubsection{Effect of the learnable template $\boldsymbol{T}$ \label{subsec:analysis-template}}

The template $\boldsymbol{T}$  is a learnable variable, it is optimized in a recognition oriented manner and can be located to positions beneficial to recognition. We demonstrate the superiority of the learnable template in the following two experimental setting. First, we compare its performance with its counterpart, a fixed template, denoted as $\boldsymbol{T_{fix}}$. We obtain $\boldsymbol{T_{fix}}$ using the average coordinates of the landmarks of all frontal face images (yaw angle lies in $-15^\circ \sim 15^\circ$) in VGGFace2 dataset.

The recognition performance is shown in Table~\ref{tab:different template}. We can find that higher accuracy can be gained with learnable template $\boldsymbol{T}$ than fixed template $\boldsymbol{T_{fix}}$. 

Next, we use the learned locations of $\boldsymbol{T}$ as a new fixed template, denoted as $\boldsymbol{T_{fix}}^{\dagger}$. The comparison results are also reported in Table~\ref{tab:different template}. We can see close performance is obtained using $\boldsymbol{T_{fix}}^{\dagger}$ compared with $\boldsymbol{T}$. This gives a further verification that the superiority of learnable template lies on its ability to learn a suitable deformation template for recognition.

We show the fixed/learnable template in Fig.~\ref{fig:diff-template}. It is obvious that the locations of landmarks in the learnable template is relatively loose than the fixed template. 
Intuitively, we think this may be that face with loose landmarks can make the extracted features of recognition model more distinguishable.

\vspace{-2mm}

\subsection{Comparison with state-of-the-art methods}

\begin{table}[t!]
\setlength{\belowcaptionskip}{-4mm} 
\centering{}%
\begin{tabular}{ccccc}
\hline 
\multirow{1}{*}{{\footnotesize{}Method}} & {\footnotesize{}Training Data} & {\footnotesize{}LFW} & {\footnotesize{}YTF} & {\footnotesize{}MegaFace}\tabularnewline
\hline 
\hline 
{\footnotesize{}DeepFace~\cite{deepface}} & {\footnotesize{}4.4M} & {\footnotesize{}97.35} & {\footnotesize{}91.4} & {\footnotesize{}-}\tabularnewline
{\footnotesize{}VGGFace~\cite{parkhi2015deep}} & {\footnotesize{}2.6M} & {\footnotesize{}99.13} & {\footnotesize{}97.4} & {\footnotesize{}-}\tabularnewline
{\footnotesize{}FaceNet~\cite{schroff2015facenet}} & {\footnotesize{}200M} & {\footnotesize{}99.64} & {\footnotesize{}95.1} & {\footnotesize{}-}\tabularnewline
{\footnotesize{}DeepID2+~\cite{sun2015deeply}} & {\footnotesize{}0.3M} & {\footnotesize{}99.47} & {\footnotesize{}93.2} & {\footnotesize{}-}\tabularnewline
{\footnotesize{}SphereFace~\cite{liu2017sphereface}} & {\footnotesize{}0.5M} & {\footnotesize{}99.42} & {\footnotesize{}95.0} & {\footnotesize{}97.43$^{\dagger}$}\tabularnewline
{\footnotesize{}CosFace~\cite{wang2018cosface}} & {\footnotesize{}5M} & {\footnotesize{}99.73} & {\footnotesize{}97.6} & {\footnotesize{}97.91$^{\dagger}$}\tabularnewline
{\footnotesize{}ArcFace~\cite{deng2019arcface}} & {\footnotesize{}5.8M} & {\footnotesize{}99.82} & {\textbf{\footnotesize{}98.0}} & {\footnotesize{}98.35}\tabularnewline
\hline 
{\footnotesize{}ReST~\cite{wu2017recursive}} & {\footnotesize{}0.46M} & {\footnotesize{}99.08} & {\footnotesize{}94.7} & {\footnotesize{}-}\tabularnewline
{\footnotesize{}GridFace~\cite{zhou2018gridface}} & {\footnotesize{}3.6M} & {\footnotesize{}99.68} & {\footnotesize{}95.2} & {\footnotesize{}-}\tabularnewline
{\footnotesize{}APA~\cite{an2019apa}} & {\footnotesize{}3.3M} & {\footnotesize{}99.68} & {\footnotesize{}-} & {\footnotesize{}-}\tabularnewline
\hline 

{\footnotesize{}baseline} & {\footnotesize{}3.6M} & {\footnotesize{}99.82} & {\footnotesize{}96.53} & {\footnotesize{}98.45}\tabularnewline\
{\footnotesize{}our approach} & {\footnotesize{}3.6M} & {\textbf{\footnotesize{}99.85}} & {\footnotesize{}96.63} & {\textbf{\footnotesize{}98.62}}\tabularnewline
\hline 
\end{tabular}\vspace{2mm}
\caption{\label{tab:public-benchmark}Evaluation on public benchmarks. We report
mAC on LFW and YTF, and Rank-1 on MegaFace. $^{\dagger}$ means the
reported results in ArcFace~\cite{deng2019arcface}.}
\vspace{-1mm}
\end{table}

To further verify the effectiveness of our method, we compare the performance with several state-of-the-art face recognition methods on three public benchmarks
 In this part, we use LResnet100E-IR (the same as ArcFace~\cite{deng2019arcface}) as our recognition model and it is trained on
MS-Celeb-1M dataset~\cite{guo2016ms}. Besides MegaFace~\cite{kemelmacher2016megaface},
we evaluate our method on LFW~\cite{huang2008labeled} and YTF~\cite{wolf2011face}
datasets. Among all methods,
ReST~\cite{wu2017recursive}, GridFace~\cite{zhou2018gridface}
and APA~\cite{an2019apa} are most closely related with our method
as they are all designed to enhance the face alignment.

The result in Table~\ref{tab:public-benchmark} shows that our model can achieve comparable result with the state-of-the-art methods CosFace~\cite{wang2018cosface} and ArcFace~\cite{deng2019arcface}. The performance of our model is also better than all face alignment methods. It is also noticeable that our proposed techniques, i.e. alignment supervision loss and warping on feature map, helps improve the model's performance on LFW and YTF. These results verify the generality of our method.

\vspace{-1mm}
\section{Conclusion}
In this paper, we provide a quantitative study of how alignment strength affects recognition. We find moderate alignment can boost the recognition accuracy, but excessive alignment is harmful. To strike the balance of face alignment, we propose two simple but effective techniques. First is to explicitly supervise the deformation with a novel alignment loss. By tuning the loss ratio, we can significantly improve the recognition performance. Second is to perform alignment on feature maps, instead of input images. We verify the effectiveness of our approach on several benchmarks, especially the challenging situations such as large pose.

%
%
\bibliographystyle{splncs04}
\bibliography{egbib}
\end{document}